\documentclass[11pt]{article}

\usepackage{microtype}
\usepackage{authblk}
\usepackage[utf8]{inputenc}
\usepackage[T1]{fontenc}
\usepackage{amsmath, amssymb, amsthm}
\usepackage{graphicx}
\usepackage{hyperref}
\usepackage[margin=1in]{geometry}

\usepackage{pifont}
\newcommand{\cmark}{\ding{51}}
\newcommand{\xmark}{\ding{55}}
\usepackage{algorithm}
\usepackage{algpseudocode}
\usepackage{booktabs}
\usepackage[numbers,sort&compress]{natbib}

\theoremstyle{remark}
\usepackage{algorithm}
\usepackage{algpseudocode}

\providecommand{\keywords}[1]{%
  \par\vspace{0.5em}\noindent\textbf{Keywords: }#1\par
}

\title{Toward High-Fidelity 3D Point-Cloud Learning for Brain Folding Morphology Prediction Using Trans-Unet}

\author[1]{Geran Zhao\textsuperscript{\dag}}
\author[2]{Xiaotian Li\textsuperscript{\dag}}

\author[3]{Poorya Chavoshnejad}
\author[3]{Mir Jalil Razavi}
\author[3]{Akbar Solhtalab}

\author[2]{Lijun Yin}
\author[1]{Guifang Fu\textsuperscript{*}}

\affil[1]{%
Department of Mathematics and Statistics, Binghamton University,
Vestal, NY 13902, USA}
\affil[2]{%
School of Computing, Binghamton University, Vestal, NY 13850, USA}
\affil[3]{%
 Department of Mechanical Engineering, Binghamton University, Vestal, NY 13850, USA}

\date{}
\begin{document}
\maketitle
{
  \renewcommand{\thefootnote}{\fnsymbol{footnote}}
  \footnotetext[1]{Corresponding author. Email address: \texttt{gfu@binghamton.edu}.}
  \footnotetext[2]{These two authors contributed equally.}
}

\begin{abstract}
Learning high-fidelity point-cloud features in the 3D space poses significant challenges, including permutation invariance, lack of local context, difficulty in fine-grained surface reconstruction, and high computational cost. In this article, we propose Trans-Unet, a novel framework that addresses these issues by first tansforming 3D point-cloud data into a 2D grid domain and then employing a U-shaped hybrid model that integrates Convolutional Neural Networks, and self-attention mechanisms. The proposed Trans-Unet effectively learns and reconstructs precise features from high-resolution 3D point-cloud data (with 40,401 points in surface and 2,382 points in fiber) derived from a predefined finite element brain patch growth model, enabling accurate prediction of brain folding patterns. By combining multiple techniques, Trans-Unet leverages the complementary strengths: the 3D-to-2D transformation preserves fine-grained structural information while significantly reducing computational cost and the curse of dimensionality; convolutional blocks capture hierarchical, low-level local representations; and the self-attention mechanism models global, high-level semantics and long-range dependencies. The dataset consists of 3D point-clouds containing both brain surface patches and fiber information generated by a large-scale finite element  model. Trans-Unet is applied to predict brain surface folding from the initial state (state 0 or states 0-2) to the final state (state 3). Experimental results demonstrate that Trans-Unet achieves high-resolution predictions of brain patch growth, surpassing existing methods in both fidelity and accuracy.
\end{abstract}


\section{Introduction}
During human brain development, the cerebral cortex undergoes a complex growth process in both volume and surface area, giving rise to gyri (convex ridges) and sulci (concave grooves) folding patterns \citep{vasung2016quantitative}. Primary folds emerge first, followed by secondary and tertiary folds. 
The folding patterns of the cerebral cortex vary considerably across individuals, and the size and morphology of these folds exert a significant influence on brain function. Research has also shown that these folding patterns can predict brain's cytoarchitecture \citep{fischl2008cortical}. Consequently, a deeper understanding of cortical folding is crucial for the early identification of cognitive impairments, as well as neurodevelopmental and psychiatric disorders.

However, modeling the underlying mechanisms of the brain folding process and predicting them has remained a significant research challenge. Over the past decades, scientists have suggested that brain folding arises from a complex interplay of biological and mechanical processes. Early theories emphasized external forces, such as cranial constraints and cerebrospinal fluid pressure, as the primary drivers of cortical folding \citep{jones2012cerebral}. However, recent experimental studies indicate that internal forces play a more dominant role. Differential growth within the cerebral cortex \citep{richman1975mechanical}, particularly in the outer layers that grow faster than the inner layers, has been proposed as an important mechanism of cortical folding. Although the tangential differential growth hypothesis is supported by experimental and computational studies \citep{ronan2014differential,jalil2015cortical, tallinen2016growth,razavi2018genomic}, other factors such as axon fiber positioning and growth also play an important role in shaping cortical folds  \citep{holland2015emerging,zhang2016mechanism,razavi2017radial,solhtalab2025stress}.

To understand human cerebral cortex's folding patterns, various descriptors for brain folding patterns have been proposed to quantify cortical folding, ranging from vertex curvature and small surface patches to gyri and sulcus, cortical lobes, and the entire cortical surface \citep{thompson1996surface,van1998functional,woods1998automated,fischl1999cortical,fischl2002whole,mangin2004coordinate,li2008novel,yang2008diffusion,awate2009multivariate,boucher2009oriented,li2010gyral}. Multiple hypothesis tests have also been conducted to evaluate how factors such as cortical thickness, material properties, and axonal fiber distribution influence brain folding morphology \citep{toro2005morphogenetic,tallinen2014gyrification,razavi2015role,tallinen2015mechanics,holland2015emerging,razavi2017radial,chavoshnejad2021role,garcia2021model}. For example, \citet{jalil2015cortical} have shown that faster cortical growth increases gyrification, while greater cortical stiffness or thickness reduces it. Even small variations in geometry and boundary conditions can influence brain's final structure and function, highlighting the significance and uniqueness of cortical morphology across individuals \citep{naidich2012imaging}.

Computational models have become a valuable tool for studying the complex process of cortical folding, particularly in the absence of precise, real-time, and long-term observational data on brain growth \citep{fernandez2016cerebral,wang2019early,da2021role,garcia2021model,zarzor2021two,darayi2022computational}. These models can be broadly categorized into two main types: mechanistic models and machine learning approaches. Among the mechanistic models, the finite element (FE) approach has received considerable attention \citep{bakhaty2017consistent, madani2019bridging,barbeito2020modeling}. It has been widely used to simulate brain development by incorporating prior knowledge of tissue properties and mechanical behaviors. The FE approach approximates the underlying true functions using high-resolution discrete elements: the domain is divided into numerous small elements connected through nodes, where function values are assigned. Each element provides a local, piecewise approximation defined by its nodal values \citep{courant1994variational,clough1960finite}. While increasing the number of elements can significantly improve accuracy, it also leads to a substantial rise in computational cost.
To address these computational challenges and complement the limitations of mechanistic models, researchers have increasingly turned to data-driven machine learning approaches. Machine learning techniques have been widely applied to predict brain cortical folding patterns, achieving high accuracy while substantially reducing computational costs \citep{shim2020rapid,wu2020network,menichetti2021machine}. However, machine learning models typically require large datasets for effective training. To overcome this limitation, integrating machine learning with FE methods has emerged as a promising strategy, leveraging the predictive efficiency of machine learning alongside the feasible sample requirements and interpretability of FE modeling \citep{martin2016machine,martinez2017finite,karami2018machine,yang2019modeling,madani2019bridging,hashemi2020novel,menichetti2021machine,calka2021machine,suwardi2022machine,chavoshnejad2023integrated}. 

In this article, we further advance the integration of finite element with deep neural network to improve predictive precision \citep{fleps2022review}. We propose Trans-Unet, a novel framework, by first projecting 3D point-cloud data into 2D grid domain and then applying a U-shaped hybrid model that combines Convolutional Neural Network (CNN), and self-attention mechanism. Trans-Unet effectively balances translation invariance, local pattern representation, and global context awareness in 3D point-cloud learning. Unlike conventional transformer-based point-cloud models \citep{guo2021pct,lu20223dctn}, Trans-Unet employs a bidirectional mapping between 3D and 2D. Mapping 3D data to 2D before feeding data into a deep learning model offers several advantages, including dimension reduction to avoid the curse of dimensionality, improved computational efficiency, reduced sensitivity to noise, and enhanced locality and translation invariance. However, this mapping introduces unavoidable challenges caused by non-unique points, where multiple points share the same $(x,y)$ coordinates but differ in their depth $z$ values due to occlusions in 3D space, primarily occurring during the folding process. Mapping these ambiguous points can lead to a certain level of inconsistency in proximity relationships within the resulting 2D representation. To address these issues, we incorporate a self-attention mechanism that preserves depth and spatial relationships in the 2D projection, and further combine transformer blocks with CNN layers to leverage their complementary strengths: convolutional layers excel at capturing local features in grid-structured data, while transformers effectively model global and long-range dependencies.

We apply Trans-Unet to high-resolution 3D point-cloud data generated by a FE model to predict the folding morphology of the fetal brain from 25 weeks (stage 0) to 36 weeks (stage 3), validating its accuracy in capturing the progressive development of cortical folding patterns. The input data from the FE model comprises approximately 40,401 brain surface points and 2,382 axonal fiber points, enabling a highly detailed spatial representation that is essential for understanding complex brain folding structures. This capability of processing 40,401 points substantially exceeds that of most existing point-cloud models in the literature \citep{chang2015shapenetinformationrich3dmodel,wu20153dshapenetsdeeprepresentation,pan2021variational}.

Point-cloud data pose several challenges, including permutation invariance, point-to-point interactions, and invariance under geometric transformations \citep{qi2017pointnet,ahmed2018survey}. Many approaches are built around how these constraints are handled. PointNet and PointNet++ \citep{qi2017pointnet,qi2017pointnet++} address permutation invariance with symmetric pooling while directly consuming raw points, offering an elegant baseline that remains widely adopted because of its simplicity, interpretability, and reproducibility. Yet their global aggregation tends to under-represent fine local structures and becomes sensitive to noise at higher resolutions. Alternative methods attempted richer local modeling through neighborhood-aware operations. Convolution-inspired networks such as SpiderCNN, PointCNN, and PointConv introduce learnable kernels or $\chi$-conv transformations to aggregate spatial context, which improves geometric fidelity but increases computational cost and suffers from overfitting on small datasets \citep{xu2018spidercnndeeplearningpoint,li2018pointcnn,wu2020network}. Attention-driven designs, exemplified by the point-cloud Transformer (PCT) \citep{guo2021pct}, embed self-attention to capture long-range interactions, showing strong segmentation accuracy on moderate-scale benchmarks but still facing scalability and robustness challenges for high-resolution scans.

A complementary line of work sidesteps raw-point irregularity by mapping 3D point-cloud into structured 2D representations. Multi-view projection techniques \citep{bourbia2022blind} transfer 3D objects to 2D from multiple viewpoints and process the multiple resulting 2D images leveraging mature image backbones to extract semantically rich features. While this approach preserves some geometric information, it does not fully retain exact 3D point positions. Additionally, multi-view projection incurs computational overhead during rendering and poses challenges in fusing information from multiple perspectives into a coherent 3D representation.

In summary, the proposed Trans-Unet partially reduces the difficulty of permutation invariance through only one top-view 3D-to-2D mapping, while still maintaining partial perturbation invariance. Directly processing tens of thousands of unordered points forces PointNet-style encoders to rely on aggressive subsampling or heavy pooling, which can dilute fine-scale geometry. Multi-view approaches partially address this but require rendering multiple high-resolution views, which is computationally expensive and introduces view-fusion artifacts. In contrast, Tran-Unet projects all points into a single 2D representation using a partially flexible mapping that preserves essential geometric context. This approach avoids the overhead and fusion complexity of multi-view methods while retaining sufficient structure to recover discriminative features. A CNN backbone efficiently captures local neighborhood patterns, while a transformer block considers the global context to correct residual irregularities introduced during the 3D-to-2D mapping. By balancing efficiency, fidelity, and invariance handling, Tran-Unet is particularly suited for high-resolution point-cloud prediction, where traditional point-based models require aggressive subsampling that can compromise geometric detail. We therefore evaluate our method against PointNet and PointNet++, which provide well-established reference points for assessing the benefits of the proposed approach.

\section{Methodology}

In this section, we introduce the proposed Trans-Unet from its five key components: bidirectional 3D-to-2D mapping, CNN-based network with a symmetric function, Transformer module with multi-head Self-Attention, Unet style encoding-decoding framework, and data augmentation skill to effectively increase the number of training samples.

\subsection{Bidirectional 3D-to-2D Mapping}
\label{sec:mapping} 
Our 3D-to-2D mapping idea draws inspiration from UV mapping, a technique that projects a 3D surface onto a 2D plane \citep{article}. This approach is commonly employed in 3D face reconstruction \citep{zhu2017face,8578839,Na_2020}. The key motivation for adopting UV mapping is its ability to impose a consistent 2D parameterization on unordered 3D point clouds, thereby mitigating the permutation invariance and curse of dimensionality challenges in later deep learning process. 

During the 3D-to-2D mapping process, two scenarios require different handling. First, a unique point, where each $(x,y)$ coordinate value corresponds to exactly one $z$ value from the 3D point-cloud. Second, a non-unique point, where multiple points share the same $(x,y)$ coordinate value but differ in $z$ values, typically due to occlusions in 3D space. The bidirectional mapping algorithm is presented in Algorithm \ref{alg:mapping}.

Let $N$ denote the total number of points in the 3D point-cloud, represented in XYZ coordinates, where $N$ is the square of an integer in the given dataset. Let $i \in \{ 0, 1, \dots, N-1\}$ stands for the index of each of the point $P_i$. Define the 3D point-cloud as
\[
\mathcal{P}_{3D} = \{ (x_i, y_i, z_i) \in [-1, 1]^3 \mid i = 0, \dots, N-1 \},
\]
where each point lies within the normalized coordinate range $[-1, 1]$ along all three axes. After mapping to a 2D grid domain, let \((u_i, v_i)\) denote the 2D grid indices in the UV domain, which is defined as
\[
\mathcal{P}_{2D} = \{ (u_i, v_i) \in \{ 0, 1, \dots, \sqrt{N}-1 \} \times \{ 0, 1, \dots, \sqrt{N}-1 \} \mid i = 0, \dots, N-1 \}.
\]

\subsubsection{Two Different Mapping Schemes}
We introduce a binary flag variable \( D_i \) for each 3D point \( P_i \), where \( i \in \{0, 1, \dots, N-1\} \), to indicate its mapping status:
\[
D_i =
\begin{cases}
\text{done}, & \text{if the mapping for } P_i \text{ is complete}, \\
\text{void}, & \text{otherwise}.
\end{cases}
\]

Similarly, the occupancy status of a 2D grid \((u_i, v_i)\) is defined as:
\[
O_{u_i,v_i} =
\begin{cases}
\text{occupied}, & \text{if a point is assigned to } (u_i,v_i), \\
\text{empty}, & \text{otherwise by default}.
\end{cases}
\]
For each unique 3D point \((x_i, y_i, z_i)\), we apply the following strict mapping formula to assign it to a corresponding 2D grid index \((u_i, v_i)\), where the indices are scaled as:
\begin{equation}
u_i = \left\lfloor \frac{x_i - x_{\min}}{x_{\max} - x_{\min}} \cdot (\sqrt{N}-1) \right\rceil, \quad
v_i = \left\lfloor \frac{y_i - y_{\min}}{y_{\max} - y_{\min}} \cdot (\sqrt{N}-1) \right\rceil,
\label{uvformula}
\end{equation}
where \(\lfloor \cdot \rceil\) denotes rounding to the nearest integer, and \(x_{\min}, x_{\max}, y_{\min}, y_{\max}\) are the minimum and maximum values of the 3D coordinates along the $x$ and $y$ axis. After mapping, their $D_i$'s are labeled as \{\text{done}\} and their corresponding UV grid locations $O_{u_i,v_i}$'s are labeled as \{\text{occupied}\}. 

For non-unique 3D points that share the same \((x,y)\) coordinates, we first group them by their common \((x,y)\) value and define \(\mathcal{S}_{(x,y)} = \{(x_i, y_i, z_i) \in \mathcal{P}_{3D} \,\mid\, (x_i, y_i) = (x, y)\}\) be the set of points sharing the same \((x, y)\) value. Then we sort all points within each \(\mathcal{S}_{(x,y)}\) in descending order of their \(z\)-values. The point with the largest \(z\)-value in each \(\mathcal{S}_{(x,y)}\) set is mapped to the UV domain following the strict formula (\ref{uvformula}). The remaining points in each \(\mathcal{S}_{(x,y)}\) set are labeled as \(D_i = \{\textit{Void}\}\) for subsequent processing. For example, consider three points \(P_0=(x,y,z_0)\), \(P_1=(x,y,z_1)\), and \(P_2=(x,y,z_2)\) that share the same \((x,y)\) coordinates but differ in \(z\). After sorting these three points by their \(z\)-values, assume \(z_1\) is the largest. Then, \(P_1\) is mapped using the strict mapping formula (\ref{uvformula}). Meanwhile, we assign \(D_1=\{\textit{done}\}\), \(D_0=\{\textit{Void}\}\), and \(D_2=\{\textit{Void}\}\) to indicate that further mapping is required for \(P_0\) and \(P_2\). We also update the occupancy status of the corresponding UV status of $P_1$ as \(O_{u_1,v_1}=\{\textit{occupied}\}\). Finally, all points labeled as \{\textit{Void}\} (e.g., \(P_0\) and \(P_2\)) are processed by mapping them to the nearest unoccupied UV grid locations using a nearest-neighbor search. 

Since each point \( P_i \) is associated with a UV coordinate \((u_i, v_i)\) and its corresponding 3D coordinate \((x_i, y_i, z_i)\), this establishes a unique one-to-one mapping between the 3D space and the 2D UV domain. Consequently, the original 3D coordinates can be accurately recovered from the UV representation whenever needed:
\[
P_i \longleftrightarrow (x_i, y_i, z_i) \longleftrightarrow (u_i, v_i) .
\]
After this UV mapping, the 3D point-cloud with \( N = 40{,}401 \) points is arranged into a \( 224 \times 224 \) UV domain. We retain the $Z$-coordinate by replicating its value across all RGB channels for each UV grid point before feeding it into the deep learning pipeline.

\renewcommand{\algorithmicdo}{}
\renewcommand{\algorithmicthen}{}%

\begin{algorithm}
\caption{The Bidirectional 3D-to-2D Mapping}
\label{alg:mapping}
\begin{algorithmic}[1]

\Procedure{Map3DTo2D}{}
    \State Let $\mathcal{P}_{3D} = \{ (x_i, y_i, z_i) \in [-1, 1]^3 \mid i = 0, \dots, N-1 \}$ be the set of $N$ points in 3D space.
    \If{Points are unique, where each $(x,y)$ coordinate value corresponds to exactly one $z$ value.
    \For {each of the unique points}
        \State Perform a strict mapping following the formula (\ref{uvformula}).
        \State Label $D_i$ as \{\text{done}\} and $O_{u_i,v_i}$ as \{\text{occupied}\}.
    \EndFor}
    \ElsIf{Points are non-unique, where each $(x,y)$ coordinate value corresponds to multiple different $z$ values.
        \State Group all points with the same $(x,y)$ into $\mathcal{S}_{(x,y)}$ set.
        \For {each $\mathcal{S}_{(x,y)}$ set}
       \State Sort all points in \(\mathcal{S}_{(x,y)}\) in non-increasing order of their \(z\) values.
        \State Select the point from $\mathcal{S}_{(x,y)}$ with the largest $z$ value, map it to 2D grid domain following formula (\ref{uvformula}). Label $D_i$ as \{\text{done}\} and $O_{u_i,v_i}$ as \{\text{occupied}\}.
        \State Label all remaining points in $\mathcal{S}_{(x,y)}$ as $D_i=\{\text{void}\}$.
        \EndFor
        \For {each of the void points, map it to the nearest unoccupied UV grid locations using a nearest-neighbor search.}
        \EndFor
        }
    \EndIf
\State After the UV mapping is complete, retain the $Z$-coordinate by replicating its value across all RGB channels for each UV grid point \((u_i, v_i)\).
\EndProcedure
\end{algorithmic}
\end{algorithm}

\subsubsection{Possible Pitfalls and Solutions}
The proposed 3D-to-2D mapping is model-free, computationally efficient, and preserves local geometric relationships, ensuring that most spatially adjacent points in 3D remain close in the UV domain, particularly those with high depth values near the surface. To account for points occluded behind surface, we sort points by their $z$-values and processing them gradually, ensuring that no critical information is lost during the mapping. Non-unique points, which typically represent important folding areas and are generally occluded, can shift during the nearest-neighbor step, introducing offsets in their 2D positions. This displacement can be mitigated by subsequent components of Trans-Unet, such as the transformer blocks, which improve point alignment and reduce errors for these ambiguous points. By keeping occluded points and their relative depth ordering, the 3D-to-2D mapping maintains the integrity of the mapped scene and ensuring that the 2D grid accurately reflects the underlying 3D structure.

\subsection{The U-Attention Architecture}
\label{subsection: Arch} 

\begin{figure}
    \includegraphics[width=\columnwidth]{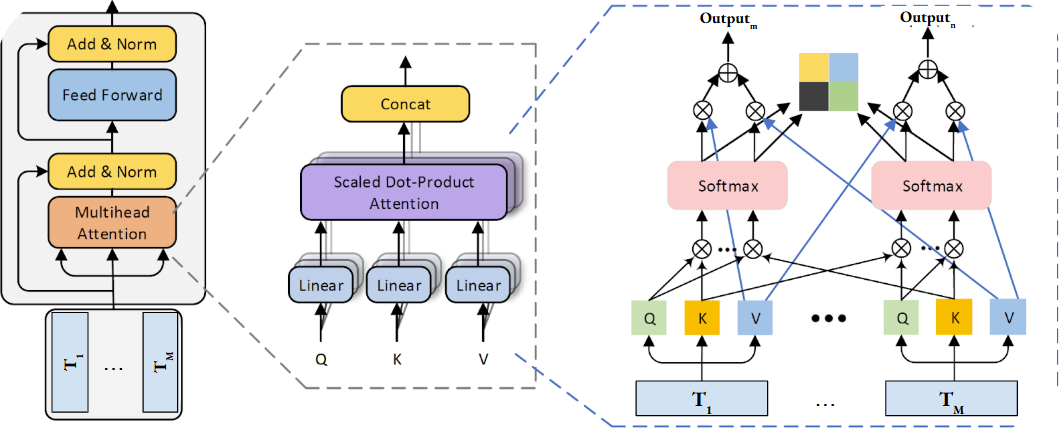}
    \caption{Transformer block}
    \label{fig: trans}
\end{figure}

After performing the 3D-to-2D mapping, we construct a series of deep neural network architectures on the resulting 2D images. As a standard approach for image data, we begin with convolutional neural networks. To mitigate potential degradation issues, we incorporate residual learning blocks \citep{he2016deep}. However, even with residual connections, CNNs face limitations in capturing the complex interrelationships among features. This challenge is further exacerbated by the characteristics of the brain surface dataset motivating this research, which is both relatively small in sample size and highly complex in structure. Consequently, CNNs alone are insufficient for effectively learning and representing the subtle patterns underlying cortical folding.

To enhance the capability of CNNs, we integrate a transformer block (see Figure \ref{fig: trans}) that employs self-attention mechanisms to capture global feature dependencies \citep{vaswani2017attention}. Notably, we omit positional embeddings in the self-attention module, which improves flexibility and tolerance to point shifts introduced during the 3D-to-2D mapping. This design enhances adaptability by reducing sensitivity to positional variations while retaining a certain level of permutation invariance. 

Specifically, after passing the transformed 2D input images through the residual blocks, a sequence of local feature representations is extracted. To model the relationships between these local features and the global context, the features are subsequently processed by a standard transformer block. The multi-head attention mechanism applies the following transformation: 
\begin{equation}
\text{MultiHead}(Q, K, V) = \text{Concat} (\text{Head}_{1},...,\text{Head}_{H})W^O,
\end{equation}
where $W^O$ is a learnable weight matrix, and $Q$, $K$, and $V$ represent a set of query, key, and value of the multi-head attention block. Then the $h$th attention matrix is calculated as:

\begin{equation}
\begin{aligned}
\text{Attention}(Q, K, V) 
   &= \text{softmax}\!\left(\frac{QK^T}{\sqrt{d_k}}\right)V, \\
\text{Head}_{h} 
   &= \text{Attention}(QW_{q,h}, KW_{k,h}, VW_{v,h}), \quad h=1,\ldots,H.
\end{aligned}
\end{equation}

where $W_{q,h},W_{k,h}$ and $W_{v,h}$ are parameter matrices corresponding to $Q, K$, and $V$, respectively, and $d_{k}$ is the dimension of key. 
Incorporating the residual connection, the output of the self-attention module is computed as $L_{Attention} =\text{LayerNorm}(\text{MultiHead}(Q,K,V)+T_m)$. The feed-forward layer consists of two linear transformations with a GELU non-linearity. Its output is added to $L_{Attention}$, followed by layer normalization. The output of transformer block is calculated, and reconstructed as a 2D latent feature.

To handle the permutation invariant property, we also induced a symmetric function \citep{qi2017pointnet} which is simple symmetric function but can also capture different properties of the dataset. The symmetric function that we employed is max-pooling.

\begin{figure*}
\hspace*{-1.8em}
\begin{center}
	\includegraphics[width=\linewidth]{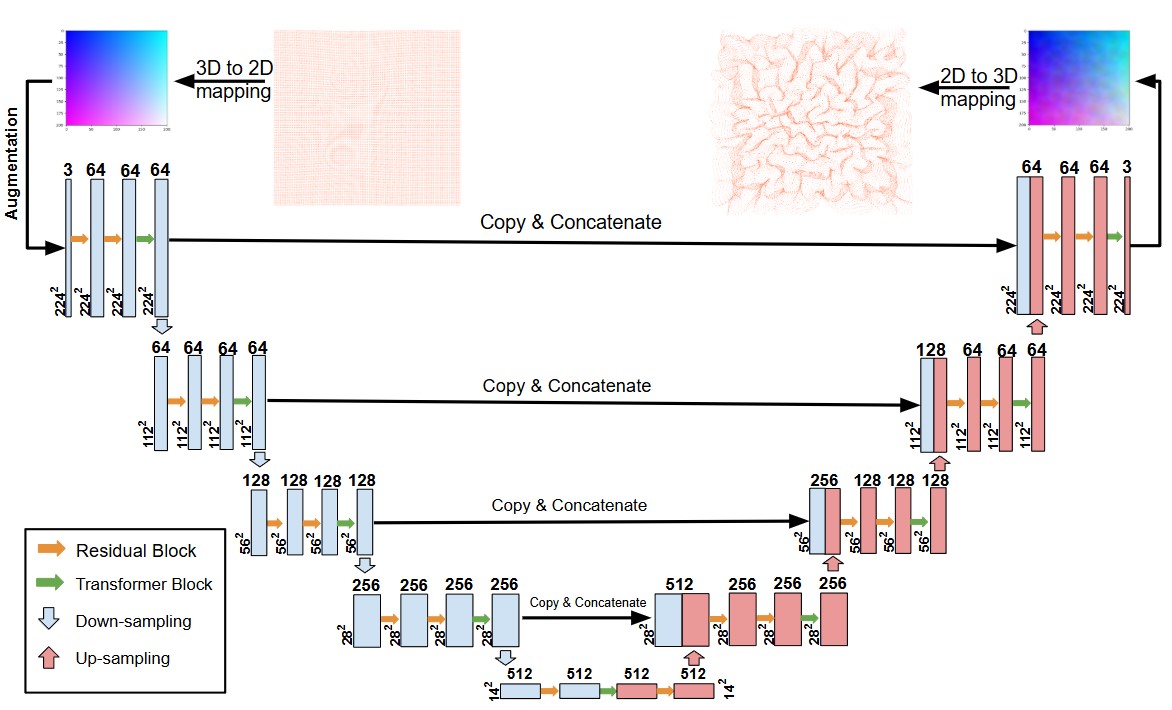}
    \caption{The U-Attention Architecture.}
    \label{fig:arch}
\end{center}
\end{figure*}

The network's backbone is U-Net \citep{ronneberger2015u}, chosen for its ability to capture multi-scale contextual information and reconstruct fine-grained spatial details from limited training data. The network encoder captures high-level features and context from the input image. Each layer in the encoder typically halves the spatial dimensions of the feature maps while increasing their depth. The decoder part upsamples the latent feature maps and combines them with corresponding feature maps from the encoder, through skip connections design, to produce the final predicted morph image. This path restores the spatial resolution of the image. Skip connections concatenate feature maps from the encoder to those in the decoder. This helps in retaining fine-grained details and spatial information that might be lost during the downsampling process in the encoder. The encoder and decoder are symmetric, which means that the number of feature maps at each level in the encoder is matched by the number of feature maps in the decoder. This symmetry helps align the spatial features during the concatenation via skip connections. Batch Group Normalization (BatchGP) is adopted in this network \citep{zhou2020batch}. Unlike standard Batch Normalization \citep{ioffe2015batch}, which normalizes across the entire batch and all feature channels, BatchGP normalizes across groups of feature channels, divides the channels into groups, and computes statistics (mean and variance) within each group. One of its main advantages is reduced dependency on batch size compared to naive Batch Normalization, making it particularly effective for small-batch training. We refer to the entire architecture as U-Attention, as illustrated in Figure~\ref{fig:arch}.

\subsection{Data Augmentation and Cross Validation}
We randomly split the dataset into training and testing sets at an 8:2 ratio. Given the limited number of training images, we apply a complex data augmentation techniques, including random rotations within $[-180^{\circ}, 180^{\circ}]$, random flipping (horizontal and vertical), and Gaussian noise injection. These augmentations help improve model generalization by introducing variations while preserving the structural integrity of the input data.

\section{Loss Function}
We design a comprehensive loss function that combines multiple components to effectively guide model training and improve the accuracy of the generated outputs. The loss function consists of a standard $L_2$ loss, a total variation (TV) loss, a latent loss derived from intermediate feature layers, and a perceptual loss measured using the Learned Perceptual Image Patch Similarity (LPIPS) metric. Before detailing each component, we first restate the important notation. Here, $G$ denotes the ground-truth 2D image (the brain folding morphology at stage 3 in this application), and $\hat{G}$ represents the predicted 2D image output from the Trans-Unet model. 

\subsection{\texorpdfstring{$L_2$}{L2} Loss}
We start with the $L_2$ loss, which is a widely used metric for training models in image prediction tasks. The $L_2$ loss measures the squared Euclidean distance between the predicted image and the ground-truth image, and is defined as:
\begin{equation}    
L_2 = \sum_{i=0}^{\sqrt{N}-1}\sum_{j=0}^{\sqrt{N}-1} (G_{i,j} - \hat{G}_{i,j})^2.
\end{equation}

\subsection{Total Variation Loss}
To enhance spatial smoothness in the generated output and mitigate high-frequency noise, we incorporate the TV loss \citep{RUDIN1992259}. This loss function takes effect as a regularization term that penalizes large intensity differences between neighboring pixels (or points, in the case of point-cloud data), thereby promoting smoother and more coherent outputs. The TV loss is defined as:

\begin{equation}
L_{TV} = \sum_{i=0}^{\sqrt{N}-1}\sum_{j=0}^{\sqrt{N}-1} \left( \left| \hat{G}_{i+1,j} - \hat{G}_{i,j} \right| + \left| \hat{G}_{i,j+1} - \hat{G}_{i,j} \right| \right),
\end{equation}
where the summation spans neighboring points in the generated output, penalizing substantial changes between them. Including TV loss encourages the model to produce spatially coherent outputs with minimal visible artifacts, particularly along object boundaries. 

\subsection{The Latent Intermediate Layer Feature-based Loss (Latent Loss)}
To encourage the model to learn structured and semantically meaningful feature representations, we incorporate a latent loss \citep{gatys2015neuralalgorithmartisticstyle}, computed from intermediate feature activations of the network. The latent loss aligns the intermediate feature embeddings of the predicted output with those of the ground-truth image. Formally, let $\phi_l(\cdot)$ denote the feature embeddings at the $l$th intermediate layer. The latent loss is then defined as the squared difference between the intermediate feature embeddings of the predicted and ground-truth images:

\begin{equation}
L_{latent} = \sum_{l \in \mathcal{L}} \left\| \phi_l(\hat{G}) - \phi_l(G) \right\|_2^2,
\end{equation}
By minimizing this loss, the model is guided to produce outputs whose intermediate feature representations capture the essential structure and visual patterns of the ground-truth image, beyond mere pixel-wise similarity. Latent loss helps maintain a consistent and well-structured latent space, improving generalization and stability during training.

\subsection{Perceptual Loss}

To ensure perceptual fidelity in the generated output, we incorporate a perceptual loss based on the LPIPS metric \citep{johnson2016perceptuallossesrealtimestyle}. LPIPS measures the perceptual similarity between two images by comparing high-level feature representations extracted from a pre-trained Visual Geometry Group (VGG) network \citep{simonyan2015deepconvolutionalnetworkslargescale}. The perceptual loss is defined as:

\begin{equation}
L_{perceptual} = \frac{1}{C_lH_lW_l}\|  \psi_l(\hat{G}) - \psi_l(G) \|_2^2,
\end{equation}
where $\psi_l(.)$ denotes the feature map extracted from the $l$-th layer of the pre-trained VGG network $\psi$. Here, $C_l$, $H_l$, and $W_l$ represent the channel, height, and width of the feature map at layer $l$, respectively. In this experiment, we set $l$ to be 12. By using LPIPS, the perceptual loss encourages the model to focus on visually important characteristics such as texture and structure rather than relying solely on pixel-wise accuracy. Leveraging the pre-trained VGG network, which is specifically designed to capture human visual perception, improves the perceptual quality of the generated output and better aligns it with human visual judgment.

\subsection{Overall Loss Function}
The overall loss function is formulated as a weighted combination of the $L_2$ loss, TV loss, latent loss, and perceptual loss:

\begin{equation}
L_{total} = \alpha L_2 + \beta L_{TV} + \gamma L_{latent} + \delta L_{perceptual},
\end{equation}
where $\alpha$, $\beta$, $\gamma$, and $\delta$ are the hyperparameters that control the relative importance of each component. In this experiments, we set $\alpha = 1.0$, $\beta = 0.1$, $\gamma = 0.1$, and $\delta = 0.3$.

This multi-term loss function ensures that the model not only preserves accurate geometric structures but also produces perceptually realistic and smooth outputs. It promotes the capture of both low-frequency structural information and high-frequency details, enabling the generation of high-quality point-cloud predictions with improved visual fidelity and spatial consistency.

\subsection{Performance Evaluation}
To quantitatively assess the prediction accuracy, we employ the Chamfer Distance (CD), a widely adopted metric for comparing two point sets \citep{fan2017point}. The CD is defined as:

\begin{equation}
\label{CD_equation}
 d_{CD}(\mathcal{P},\hat{\mathcal{P}}) = \sum_{\xi \in \mathcal{P}} \underset{\hat{\xi} \in \hat{\mathcal{P}}}{min} \left \| \xi-\hat{\xi} \right \|_2^2+\sum_{\hat{\xi} \in \hat{\mathcal{P}}} \underset{\xi \in \mathcal{P}}{min} \left \| \xi-\hat{\xi} \right \|_2^2,   
\end{equation}
where $\mathcal{P}$, $\hat{\mathcal{P}} \subseteq \mathbb{R}^3$ denote the ground truth and the predicted 3D point-clouds, respectively. The Chamfer Distance quantifies how closely two point sets align by summing the squared distances from each point in one set to its closest counterpart in the other set, computed in both directions. A smaller CD indicates that the predicted point-cloud is closer to the ground truth in terms of both geometry and spatial distribution. The metric is particularly suitable for point-cloud comparison because it does not require explicit point-to-point correspondence.

\section{Data Analysis}
\subsection{Data background}
\begin{figure*}[!t]
    \centering
    \includegraphics[width=0.85\linewidth]{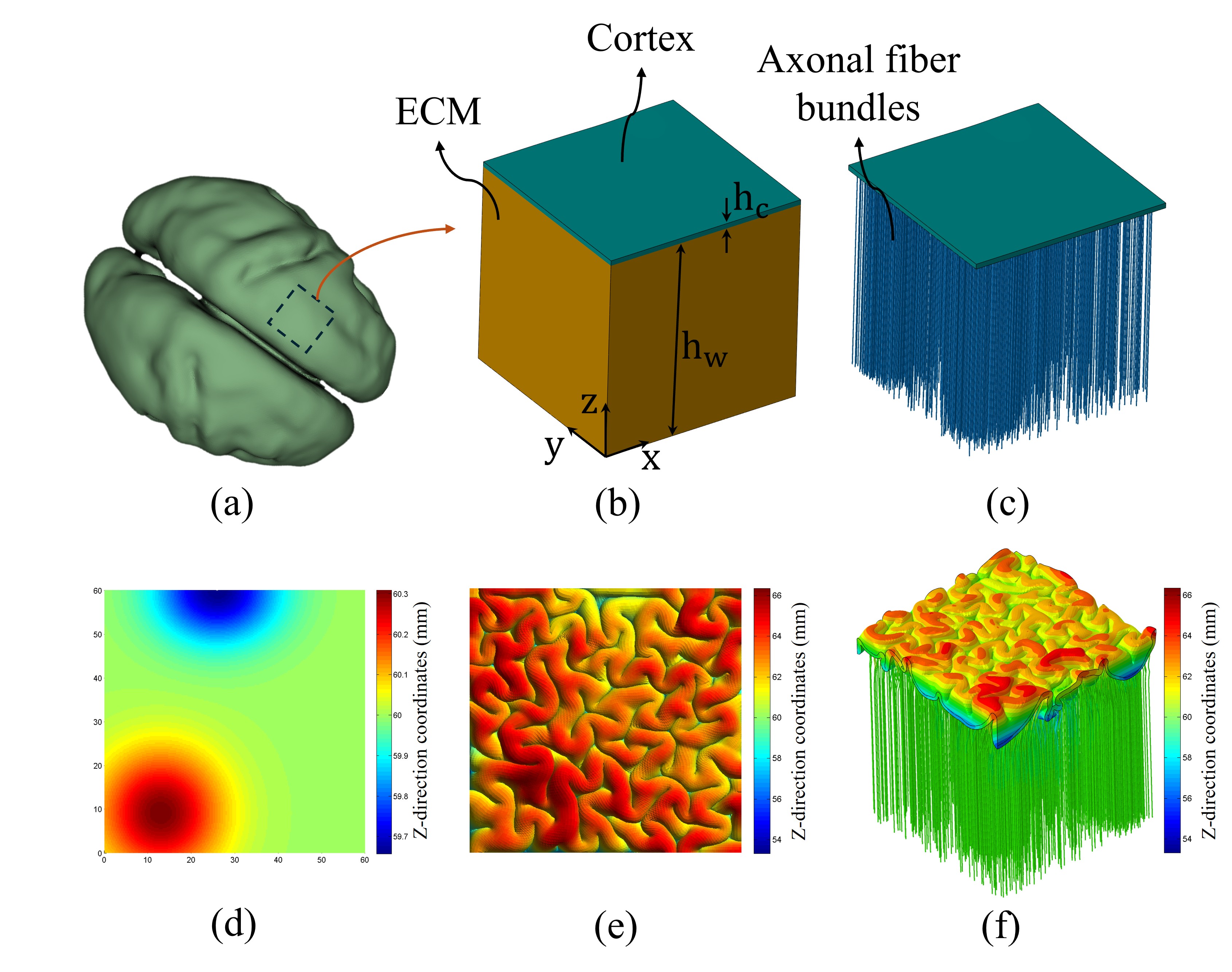}
    \caption{A patch-based growth model incorporating axonal fiber bundles to simulate brain folding. a) Reconstructed human fetal brain from MRI data. b) A representative cubic patch within the fetal brain showing the cortex surface and underlying white matter. This is the initial configuration of the 3D model used to simulate growth and cortical folding. The top layer corresponds to the cortex, and the bottom layer represents the white matter, which includes extracellular matrix (ECM) and axonal fiber bundles. The fiber bundles are embedded within the ECM and are not visible in this panel. c) Visualization of axonal fiber bundles with the ECM suppressed. Axonal fiber bundles occupy 13\% of the white matter volume. d) Thickness variation of the cortex introduced by Gaussian perturbations. The color bar represents the coordinates of nodes in the Z-direction. e) Deformed configuration of the model after simulated growth and folding. f) 3D  view of the model after growth and folding.}
    \label{fig:FEM}
\end{figure*}

A cubic model of size 60 × 60 × 60 mm was constructed to represent a localized region of the brain for investigating gyrification under cortical thickness variations (Figure \ref{fig:FEM}). The cortical layer was initially assigned a thickness of 1.5 mm, while the diameter of axonal fiber bundles was set to 500 µm \citep{chavoshnejad2021role}, providing a balance between physiological relevance and computational efficiency by controlling mesh density and fiber count. Axonal fiber bundles occupied 13\% of the white matter volume, with the remaining space filled by ECM. Four configurations were generated with 4, 5, 6, and 7 regions of high axonal fiber density. In each case, 3\% of the white matter was populated with axonal bundles using a random uniform distribution, while the remaining 10\% was distributed according to a random normal distribution centered around the high-density sites.

Both gray and white matter were discretized using 8-node linear brick elements with reduced integration (C3D8R) in ABAQUS. To optimize computational efficiency, a finer mesh was employed near the cortical surface, while a coarser mesh was used in deeper regions. A mesh convergence study was conducted to ensure that the results were independent of mesh size.

To introduce geometrical thickness variations, a custom Python script was developed to reposition nodes in the meshed model, thereby generating different initial surface morphologies. These variations were implemented using Gaussian distribution functions, which mimic the surface irregularities observed in the fetal brain and facilitate cortical folding \citep{cao2012buckling}. For each model, the number, amplitude, and position of Gaussian variations were randomly selected. Importantly, these thickness variations were applied to the cortex prior to growth simulations, ensuring that their specifications remained independent of subsequent growth. All models incorporated distinct fiber configurations and unique cortical thickness variations, as both were randomly generated. Symmetric boundary conditions were imposed along all lateral surfaces, while the bottom face was constrained in the Z-direction.

All tissue components, the cortex, ECM, and axonal fiber bundles, were modeled as nonlinear hyperelastic materials using the neo-Hookean formulation to capture large deformations associated with brain growth and folding \citep{chavoshnejad2024theoretical,solhtalab2025stress,solhtalab2025mechanics}. Axonal fiber bundles were implemented using the material embedded method: the location of each fiber bundle was specified first, and then a custom Python script mapped the material properties of the fibers onto the corresponding ECM elements. Consequently, elements containing fibers had overlapping material properties; to avoid stiffness redundancy, the ECM contribution was subtracted from the fiber elements \citep{chavoshnejad2021role}.

A self-contact constraint was applied to the cortical surface to prevent self-penetration during deformation. Cortical growth was modeled using relative tangential growth, consistent with prior studies showing that gyrification arises primarily from differential growth between gray and white matter \citep{budday2014role,tallinen2016growth}. To capture this mechanism, growth was restricted to the cortical layer, since the difference in growth rates between layers, rather than their absolute values, is the key driver of folding initiation and progression. Tangential expansion in the x- and y-directions was calibrated to produce a post-folding cortical thickness of approximately 3 mm. The growth protocol was kept identical across all models, with variations arising only from differences in their initial surface morphology.

\begin{figure}
     \centering
     \includegraphics[width=0.6\linewidth]{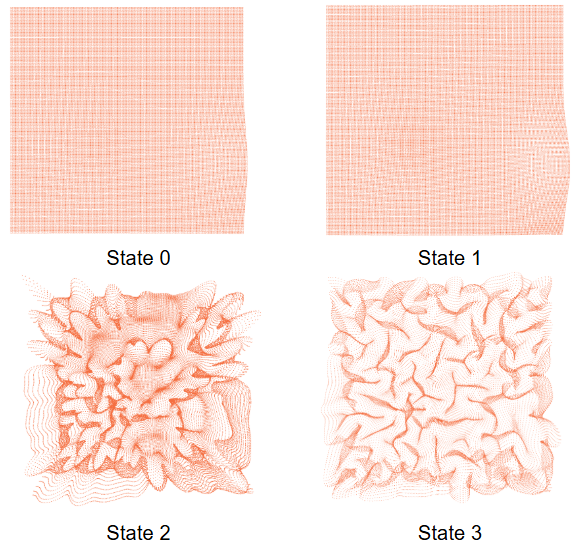}
     \caption{An example of brain surface data: from state 0 to state 3.}
     \label{dataimage}
 \end{figure}

The brain surface datasets encompass four developmental stages, illustrating the progression from a smooth cortex to a highly folded morphology (Figure \ref{dataimage} shows these stages in 3D point-cloud format).

\subsection{Comparison under eight experimental settings}
We compare Trans-Unet against two well-establish point-based approaches, PointNet and PointNet++ \citep{qi2017pointnet,qi2017pointnet++}.  To ensure a comprehensive evaluation, we assess the performance of these three methods under various experimental settings, analyzing the prediction results both quantitatively and qualitatively. Before presenting the results, we first descbribe the experimental setups.  \begin{itemize}
        \item PointNet-1: Using PointNet to predict growth progression from state 0 to state 3, with each image consisting of 40,401 XYZ points in a 3D point-cloud.
        \item PointNet++-1: Using PointNet++ to predict growth progression from state 0 to state 3, with each image consisting of 40,401 XYZ points in a 3D point-cloud.
        \item PointNet-2: Prediction by PointNet using 8,081 points. These 8,081 points are obtained through even down-sampling from the original 40,401 points.
        \item PointNet++-2: Prediction by PointNet++ with 8,081 points.
        \item Trans-Unet-1: Prediction by Trans-Unet from state 0 to state 3 with 40,401 points without fiber and augmentation.
        \item Trans-Unet-2: Prediction by Trans-Unet from state 0 to state 3 with 40,401 points and fiber but no augmentation.
        \item Trans-Unet-3: Prediction by Trans-Unet from state 0 to state 3 with 40,401 points, fiber, and augmentation.
        \item Trans-Unet-4: Prediction by Trans-Unet from states 0-2 to state 3 with 40,401 points, fiber, and augmentation.
        
    \end{itemize}
    
\begin{figure}
     \centering
     \includegraphics[width=0.7\linewidth]{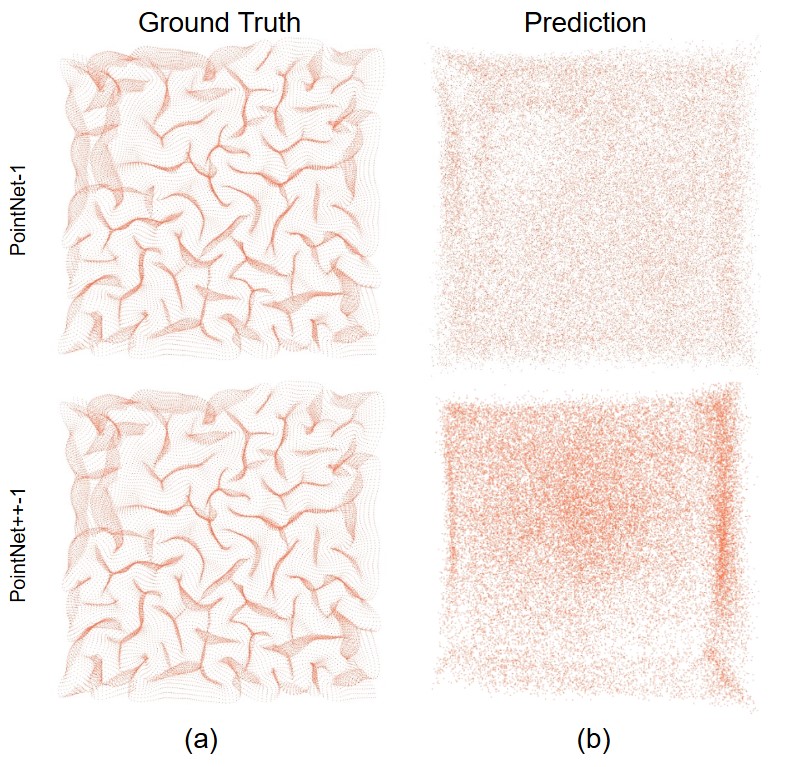}
     \caption{(a) The first column shows the ground-truth 3D point-cloud at state 3 with 40,401 points. (b) The second column shows the predicted 3D point-cloud at state 3 generated by the point-based models. The first row is PointNet-1 setting and the second row is PointNet++-1 setting.}
     \label{fig:pnresult1}
 \end{figure}

\begin{figure}
 \centering
 \includegraphics[width=0.65\linewidth]{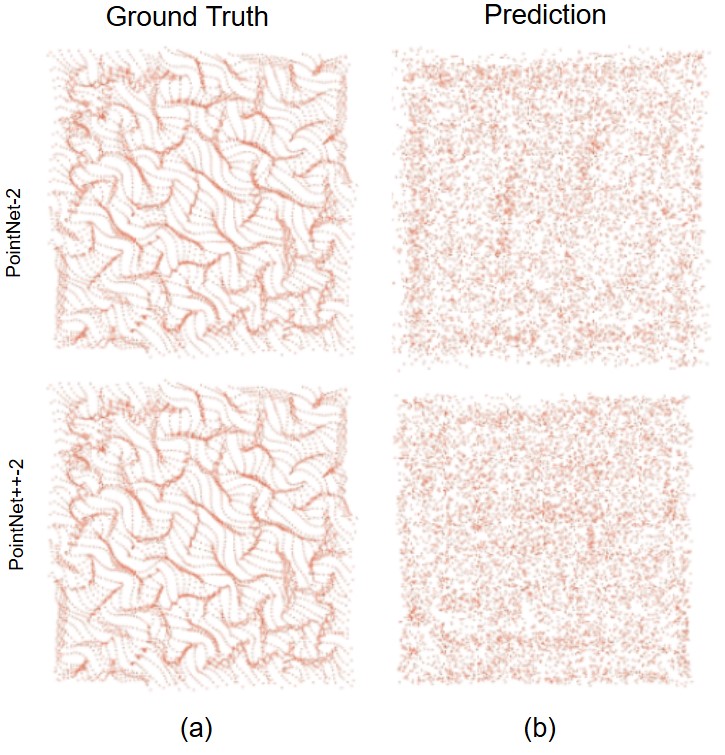}
 \caption{(a) The first column shows the ground-truth 3D point-cloud at state 3 with 8,081 points. (b) The second column shows the predicted 3D point-cloud at state 3 generated by the point-based models. The first row is PointNet-2 setting and the second row is PointNet++-2 setting.}
 \label{fig:pnresult2}
\end{figure}

As shown in Figure \ref{fig:pnresult1}, the outputs from PointNet and PointNet++ resemble random white noise and fail to capture basically meaningful brain surface folding patterns from the 40,401 point input. This limitation arises from their inability to model fine-grained local structures and potential information loss introduced by max-pooling operations. Although PointNet++ improves upon PointNet by incorporating hierarchical feature extraction, it still shares common limitations of point-based methods. To reduce the burden of processing 40,401 points, we perform downsampling by retaining one out of every five points, resulting in 8,081 points for further evaluation of point-based models. Figure \ref{fig:pnresult2} presents the outputs of PointNet and PointNet++ using 8,081 points. While both methods show some improvement compared to the results with 40,401 points, the generated image still resemble random noise without revealing any clearly recognizable patterns. 

\begin{figure}
 \centering
 \includegraphics[width=0.7\linewidth]{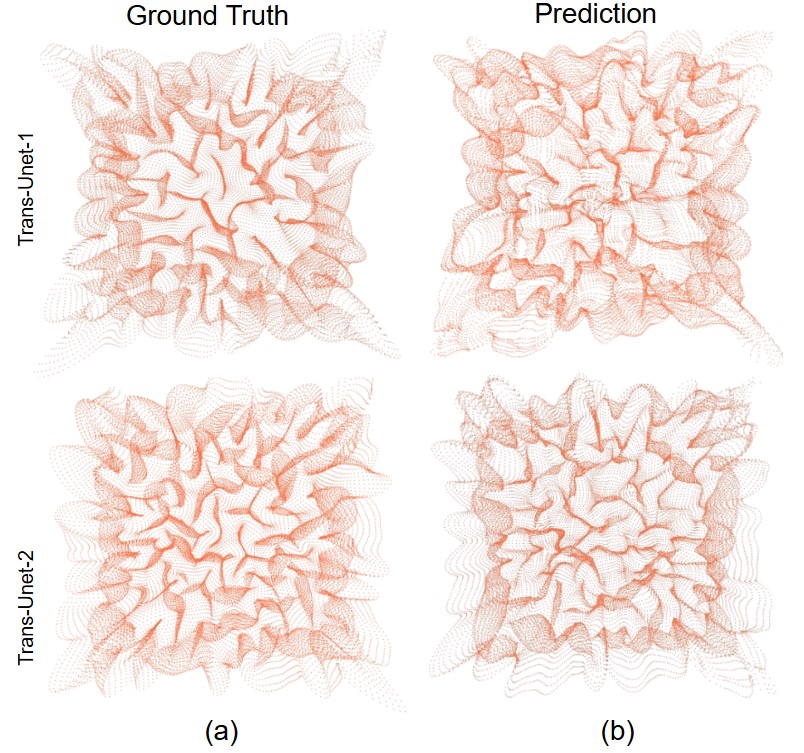}
 \caption{(a) The first column shows the ground-truth 3D point-cloud at state 3 with 40,401 points. (b) The second column shows the predicted 3D point-cloud at state 3 generated by the Trans-Unet model. The first row is Trans-Unet-1: Prediction from state 0 to state 3 without fiber and augmentation; and the second row is Trans-Unet-2: Prediction from state 0 to state 3 with fiber but no augmentation. }
 \label{fig:turesult1}
\end{figure}    

\begin{figure}
 \centering
 \includegraphics[width=0.7\linewidth]{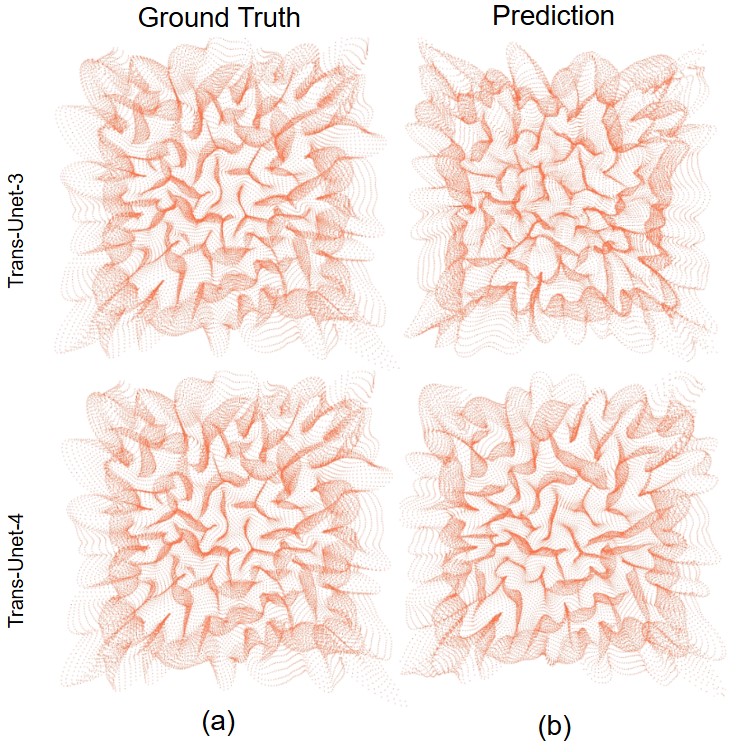}
 \caption{(a) The first column shows the ground-truth 3D point-cloud at state 3 with 40,401 points. (b) The second column shows the predicted 3D point-cloud at state 3 generated by the Trans-Unet model. The first row is Trans-Unet-3: Prediction from state 0 to state 3 with fiber and augmentation; and the second row is Trans-Unet-4: Prediction from states 0-2 to state 3 with fiber and data augmentation.}
 \label{fig:turesult2}
\end{figure}     

Compared to standard point-based methods, the proposed Trans-Unet model exhibits superior capability in accurately capturing complex brain surface folding patterns across all four experimental settings, as shown in the right columns of Figures \ref{fig:turesult1} and \ref{fig:turesult2}. Specifically, Figure \ref{fig:turesult1} compares outputs of Trans-Unet with and without fiber data, demonstrating that incorporating fiber information significantly improves prediction quality, particularly along the entire image boundaries. To further enhance performance, we investigate the effects of data augmentation and two additional input states. As expected, combining all three states (0–2), fiber data, and data augmentation yields the highest accuracy (bottom row of Figure \ref{fig:turesult2}), as later developmental states provide input signals more closely resembling the final state (state 3). Notably, the Trans-Unet variants (Trans-Unet-1, Trans-Unet-2, and Trans-Unet-3) already achieve high accuracy using only state 0 as input, despite the early folding stage being smooth, vague, and indistinct (see Figure \ref{fig:turesult1} and the top row of Figure \ref{fig:turesult2}). Figure \ref{fig:turesult2} demonstrates that Trans-Unet-4 achieves substantial improvements in reconstructing gyri and sulci. In particular, within the central region of the predicted images, the gyri and sulci in the second row appear more distinct and closely match the ground truth.

\begin{figure}
    \centering
    \includegraphics[scale=0.5]{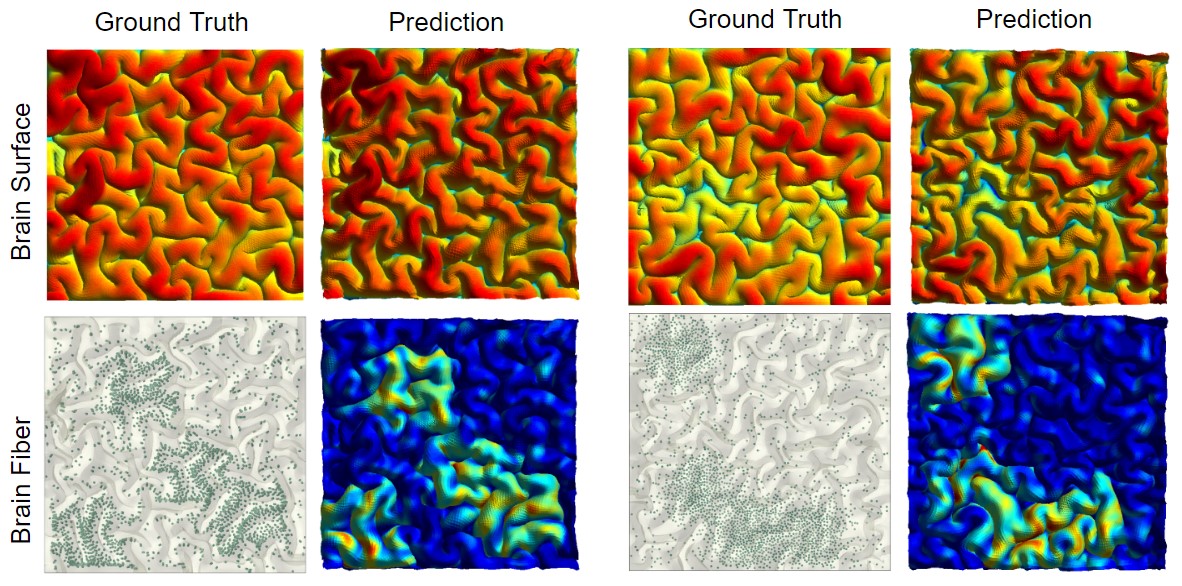}
\caption{Comparison between the ground truth and predicted results of the brain gray matter surface (top row) and fiber density (bottom row) for two randomly selected test samples. Contours are generated from the corresponding $z$ values.}
    \label{fig:result3}
\end{figure}

Figure~\ref{fig:result3} presents a case-by-case comparison of the ground truth and the Trans-Unet-4 predictions. The model demonstrates strong ability to accurately recover the positions of gyri and sulci using input from states 0–2 and fiber data. The predicted high-altitude regions (gyri) and low-altitude regions (sulci) closely match the ground truth. Moreover, we observe that regions with higher concentrations of growing axonal fibers tend to form 3-hinge gyri \citep{ge2018denser,razavi2021mechanism}.


To quantitatively evaluate the accuracy of the eight experimental settings across the three methods, we employ the Chamfer Distance. We want to highlight two fair comparisons: (i) comparing PointNet-1, PointNet++-1, and Trans-Unet-1, as these configurations all process 40,401 points without fiber data, augmentation, or additional states (only state 0 as input); and (ii) comparing Trans-Unet-1 through Trans-Unet-4, as they represent the same method under different experimental settings. As shown in Table \ref{tab:cd}, Trans-Unet-1 achieves a CD value of 0.045, which is approximately 0.0009 of that of PointNet-1 (45.2) and 0.0012 of that of PointNet++-1 (36.7), indicating a substantially closer alignment between predictions and ground truth. Although PointNet-2 and PointNet++-2 report lower CD values (both below 10), this comparison is not fair because their CD sums are computed over 8,081 rather than 40,401 points. Interestingly, incorporating augmentation and fiber data has only a subtle effect on CD (0.041 versus 0.045). As expected, Trans-Unet-4 achieves approximately one-third of the CD value of Trans-Unet-1, owing to the inclusion of multiple input states, fiber data, and data augmentation, which collectively provide richer information.

\begin{table}
\caption{The Chamfer Distance values of the eight experimental settings across the three methods.}
\resizebox{\columnwidth}{!}{%
\begin{tabular}{|l|l|l|l|l|}
\hline
Method & PointNet-1   & PointNet++-1 & PointNet-2   & PointNet++-2 \\ \hline
CD     & 45.2         & 36.7         & 9.1          & 7.2          \\ \hline
Method & Trans-Unet-1 & Trans-Unet-2 & Trans-Unet-3 & Trans-Unet-4 \\ \hline
CD     & 0.045        & 0.045        & 0.041        & \textbf{0.012}        \\ \hline
\end{tabular}%
}

\label{tab:cd}
\end{table}

\begin{table*}[t]
    \centering
    \caption{Ablation Study on Each of the Key Components}
    \label{tab:ablation_study}
    \resizebox{\textwidth}{!}{%
        \begin{tabular}{lccccc|c}
            \toprule
            \textbf{Method} & \textbf{UV Mapping} & \textbf{Trans-Unet} & \textbf{Perceptual Loss} & \textbf{Latent Loss} & \textbf{TV Loss} & \textbf{CD} \\
            \midrule
            Baseline & \cmark & \cmark & \cmark & \cmark & \cmark & \textbf{0.012} \\
            w/o 3D-to-2D Mapping & \xmark & \cmark & \cmark & \cmark & \cmark & 3.835 \\
            w/o U-Attention & \cmark & \xmark & \cmark & \cmark & \cmark & 0.043 \\
            w/o TV Loss & \cmark & \cmark & \cmark & \cmark & \xmark & 0.031 \\
            w/o Perceptual Loss & \cmark & \cmark & \xmark & \cmark & \cmark & 0.027 \\
            w/o Latent Loss & \cmark & \cmark & \cmark & \xmark & \cmark & 0.019 \\
            \bottomrule
        \end{tabular}%
    }
\end{table*}
\subsection{Ablation Study}
To assess the contribution of each key component to the overall performance of the Trans-Unet framework, we perform a systematic ablation study by selectively removing one component at a time while keeping all others unchanged. Model performance is evaluated using the Chamfer Distance. The results, summarized in Table \ref{tab:ablation_study}, provide critical insights into the significance of each component.

\subsubsection{Baseline Performance}
The complete Trans-Unet model, comprising a 3D-to-2D mapping, U-Attention architecture, perceptual loss, latent loss, and TV loss, achieves the lowest Chamfer Distance (0.012). This outcome confirms that the integrated architecture is highly effective for producing precise and high-quality predictions. The contribution of each component becomes clear when it is individually removed, in particular for those components that lead to significant performance degradation.

\subsubsection{Effect of the bidirectional 3D-to-2D Mapping}
Removing the 3D-to-2D mapping results in a dramatic increase in CD from 0.012 to 3.835, indicating a catastrophic decline in prediction accuracy. This sharp degradation underscores the critical role of 3D-to-2D mapping in preserving spatial alignment, ensuring geometric consistency, maintaining coherent feature correspondences, and addressing the permutation invariance inherent in point-cloud data. Without this structured transformation, any model will encounter great difficulty in establishing a meaningful correspondence between input and output, resulting in severe distortions in the generated predictions.

\subsubsection{Effect of U-Attention Architecture}
Eliminating the self-attention mechanism and U-shaped architecture, thus relying solely on conventional CNN operations, raises the CD from 0.012 to 0.043. This increase highlights the importance of this U-Attention architecture in refining spatial feature extraction and effectively capturing both local fine details and long-range dependencies, which are important for modeling complex structural patterns of the motivated brain folding morphology.
                                           
\subsubsection{Impact of TV Loss}
Removing TV Loss increases the Chamfer Distance to 0.031, representing a moderate decline in prediction accuracy. TV Loss functions as a regularizer by discouraging abrupt spatial variations and reducing noise. The observed performance drop indicates that TV Loss is particularly effective in suppressing high-frequency artifacts. Although its impact is less pronounced than that of 3D-to-2D mapping or the U-Attention component, the results confirm its importance in refining the final output by enhancing surface smoothness and preserving spatial coherence.

\subsubsection{Impact of Perceptual Loss}
Removing Perceptual Loss increases the Chamfer Distance to 0.027, indicating another moderate reduction in prediction accuracy. This component enforces high-level feature similarity by aligning generated outputs with ground-truth features in a pre-trained VGG feature space, thereby preserving fine details and visual coherence even when the overall geometric structure is maintained. Without Perceptual Loss, the model still produces structurally plausible outputs; however, these exhibit a noticeable loss of fine details, particularly in texture-rich regions.

\subsubsection{Impact of Latent Loss}
Removing Latent Loss increases the Chamfer Distance from 0.012 to 0.019. Although this increase is smaller compared to the ablations of other components, it still reflects a measurable decline in the performance of Trans-Unet. Latent Loss serves as a regularizer for the learned feature space, promoting well-structured and stable latent embeddings. In its absence, the model may exhibit slight inconsistencies in feature representation, which can subtly affect overall prediction accuracy.

In summary, the ablation study demonstrates that each component contributes uniquely and significantly to the overall performance of the Trans-Unet model. The lowest Chamfer Distance (0.012) is achieved when all components are integrated, underscoring the importance of their combined effect in delivering high-quality predictions.

\section{Conclusions}
In this paper, we propose Trans-Unet, a novel model that combines bidirectional 3D-to-2D mapping with a multi-head attention mechanism within a U-shaped network architecture, complemented by multiple loss functions. This design effectively addresses the challenges of permutation invariance and spatial inconsistency in point-cloud data, while enhancing the integration of biomechanical modeling and deep learning. Our framework achieves strong performance on a high-resolution dataset (as large as 40,401 points), enabling accurate and detailed predictions of brain cortical folding development. For future work, we aim to extend this approach in two directions: scaling predictions from localized patches to a complete spherical brain model, and exploring the adaptability of Trans-Unet to other point-cloud learning tasks and diverse human tissue datasets.


\bibliographystyle{plainnat}
\bibliography{reference}

@article{fischl2008cortical,
  title={Cortical folding patterns and predicting cytoarchitecture},
  author={Fischl, Bruce and Rajendran, Niranjini and Busa, Evelina and Augustinack, Jean and Hinds, Oliver and Yeo, BT Thomas and Mohlberg, Hartmut and Amunts, Katrin and Zilles, Karl},
  journal={Cerebral cortex},
  volume={18},
  number={8},
  pages={1973--1980},
  year={2008},
  publisher={Oxford University Press}
}

@article{vasung2016quantitative,
  title={Quantitative and qualitative analysis of transient fetal compartments during prenatal human brain development},
  author={Vasung, Lana and Lepage, Claude and Rado{\v{s}}, Milan and Pletikos, Mihovil and Goldman, Jennifer S and Richiardi, Jonas and Ragu{\v{z}}, Marina and Fischi-G{\'o}mez, Elda and Karama, Sherif and Huppi, Petra S and others},
  journal={Frontiers in neuroanatomy},
  volume={10},
  pages={11},
  year={2016},
  publisher={Frontiers Media SA}
}

@inproceedings{qi2017pointnet,
  title={Pointnet: Deep learning on point sets for 3d classification and segmentation},
  author={Qi, Charles R and Su, Hao and Mo, Kaichun and Guibas, Leonidas J},
  booktitle={Proceedings of the IEEE conference on computer vision and pattern recognition},
  pages={652--660},
  year={2017}
}

@article{qi2017pointnet++,
  title={Pointnet++: Deep hierarchical feature learning on point sets in a metric space},
  author={Qi, Charles Ruizhongtai and Yi, Li and Su, Hao and Guibas, Leonidas J},
  journal={Advances in neural information processing systems},
  volume={30},
  year={2017}
}

@book{jones2012cerebral,
  title={Cerebral cortex: comparative structure and evolution of cerebral cortex, part II},
  author={Jones, Edward G and Peters, Alan},
  volume={8},
  year={2012},
  publisher={Springer Science \& Business Media}
}

@article{li2018pointcnn,
  title={Pointcnn: Convolution on x-transformed points},
  author={Li, Yangyan and Bu, Rui and Sun, Mingchao and Wu, Wei and Di, Xinhan and Chen, Baoquan},
  journal={Advances in neural information processing systems},
  volume={31},
  year={2018}
}

@article{guo2021pct,
  title={Pct: Point cloud transformer},
  author={Guo, Meng-Hao and Cai, Jun-Xiong and Liu, Zheng-Ning and Mu, Tai-Jiang and Martin, Ralph R and Hu, Shi-Min},
  journal={Computational Visual Media},
  volume={7},
  pages={187--199},
  year={2021},
  publisher={Springer}
}

@article{shim2020rapid,
  title={Rapid prediction of brain injury pattern in mTBI by combining FE analysis with a machine-learning based approach},
  author={Shim, Vickie B and Holdsworth, Samantha and Champagne, Allen A and Coverdale, Nicole S and Cook, Douglas J and Lee, Tae-Rin and Wang, Alan D and Li, Shaofan and Fernandez, Justin W},
  journal={IEEE Access},
  volume={8},
  pages={179457--179465},
  year={2020},
  publisher={IEEE}
}

@article{menichetti2021machine,
  title={A machine learning approach to investigate the uncertainty of tissue-level injury metrics for cerebral contusion},
  author={Menichetti, Andrea and Bartsoen, Laura and Depreitere, Bart and Vander Sloten, Jos and Famaey, Nele},
  journal={Frontiers in bioengineering and biotechnology},
  volume={9},
  pages={714128},
  year={2021},
  publisher={Frontiers Media SA}
}

@inproceedings{karami2018machine,
  title={A machine learning approach for biomechanics-based tracking of lung tumor during external beam radiation therapy},
  author={Karami, Elham and Gaede, Stewart and Lee, Ting-Yim and Samani, Abbas},
  booktitle={Medical Imaging 2018: Image-Guided Procedures, Robotic Interventions, and Modeling},
  volume={10576},
  pages={322--328},
  year={2018},
  organization={SPIE}
}

@inproceedings{martin2016machine,
  title={Machine learning for modeling the biomechanical behavior of human soft tissue},
  author={Martin-Guerrero, Jose D and Ruperez-Moreno, Maria J and Martinez-Martinez, Francisco and Lorente-Garrido, Delia and Serrano-Lopez, Antonio J and Monserrat, Carlos and Martinez-Sanchis, Sandra and Martinez-Sober, Marcelino},
  booktitle={2016 IEEE 16th international conference on data mining workshops (ICDMW)},
  pages={247--253},
  year={2016},
  organization={IEEE}
}

@article{martinez2017finite,
  title={A finite element-based machine learning approach for modeling the mechanical behavior of the breast tissues under compression in real-time},
  author={Mart{\'\i}nez-Mart{\'\i}nez, Francisco and Rup{\'e}rez-Moreno, Mar{\'\i}a J and Mart{\'\i}nez-Sober, Marcelino and Solves-Llorens, Juan Antonio and Lorente, Delia and Serrano-L{\'o}pez, AJ and Mart{\'\i}nez-Sanchis, Sandra and Monserrat, C and Mart{\'\i}n-Guerrero, Jos{\'e} David},
  journal={Computers in biology and medicine},
  volume={90},
  pages={116--124},
  year={2017},
  publisher={Elsevier}
}

@article{calka2021machine,
  title={Machine-learning based model order reduction of a biomechanical model of the human tongue},
  author={Calka, Maxime and Perrier, Pascal and Ohayon, Jacques and Grivot-Boichon, Christelle and Rochette, Michel and Payan, Yohan},
  journal={Computer Methods and Programs in Biomedicine},
  volume={198},
  pages={105786},
  year={2021},
  publisher={Elsevier}
}

@article{wu2020network,
  title={A network-based response feature matrix as a brain injury metric},
  author={Wu, Shaoju and Zhao, Wei and Rowson, Bethany and Rowson, Steven and Ji, Songbai},
  journal={Biomechanics and modeling in mechanobiology},
  volume={19},
  pages={927--942},
  year={2020},
  publisher={Springer}
}

@article{hashemi2020novel,
  title={A novel machine learning based computational framework for homogenization of heterogeneous soft materials: application to liver tissue},
  author={Hashemi, Mohammad Saber and Baniassadi, Majid and Baghani, Mostafa and George, Daniel and Remond, Yves and Sheidaei, Azadeh},
  journal={Biomechanics and modeling in Mechanobiology},
  volume={19},
  pages={1131--1142},
  year={2020},
  publisher={Springer}
}

@article{suwardi2022machine,
  title={Machine learning-driven biomaterials evolution},
  author={Suwardi, Ady and Wang, FuKe and Xue, Kun and Han, Ming-Yong and Teo, Peili and Wang, Pei and Wang, Shijie and Liu, Ye and Ye, Enyi and Li, Zibiao and others},
  journal={Advanced Materials},
  volume={34},
  number={1},
  pages={2102703},
  year={2022},
  publisher={Wiley Online Library}
}

@article{fleps2022review,
  title={A review of CT-based fracture risk assessment with finite element modeling and machine learning},
  author={Fleps, Ingmar and Morgan, Elise F},
  journal={Current osteoporosis reports},
  volume={20},
  number={5},
  pages={309--319},
  year={2022},
  publisher={Springer}
}

@article{chavoshnejad2023integrated,
  title={An integrated finite element method and machine learning algorithm for brain morphology prediction},
  author={Chavoshnejad, Poorya and Chen, Liangjun and Yu, Xiaowei and Hou, Jixin and Filla, Nicholas and Zhu, Dajiang and Liu, Tianming and Li, Gang and Razavi, Mir Jalil and Wang, Xianqiao},
  journal={Cerebral Cortex},
  volume={33},
  number={15},
  pages={9354--9366},
  year={2023},
  publisher={Oxford University Press}
}

@article{jalil2015cortical,
  title={Cortical folding pattern and its consistency induced by biological growth},
  author={Razavi, Mir  Jalil and Zhang, Tuo and Liu, Tianming and Wang, Xianqiao},
  journal={Scientific reports},
  volume={5},
  number={1},
  pages={14477},
  year={2015},
  publisher={Nature Publishing Group UK London}
}

@book{naidich2012imaging,
  title={Imaging of the brain E-book: Expert radiology series},
  author={Naidich, Thomas P and Castillo, Mauricio and Cha, Soonmee and Smirniotopoulos, James G},
  year={2012},
  publisher={Elsevier Health Sciences}
}

@article{courant1994variational,
  title={Variational methods for the solution of problems of equilibrium and vibrations},
  author={Courant, Richard and others},
  journal={Lecture notes in pure and applied mathematics},
  pages={1--1},
  year={1994},
  publisher={MARCEL DEKKER AG}
}

@article{clough1960finite,
  title={The finite element in plane stress analysis},
  author={Clough, Ray W},
  journal={Proc. 2\^{}< nd> ASCE Confer. On Electric Computation, 1960},
  year={1960}
}

@inproceedings{ronneberger2015u,
  title={U-net: Convolutional networks for biomedical image segmentation},
  author={Ronneberger, Olaf and Fischer, Philipp and Brox, Thomas},
  booktitle={Medical image computing and computer-assisted intervention--MICCAI 2015: 18th international conference, Munich, Germany, October 5-9, 2015, proceedings, part III 18},
  pages={234--241},
  year={2015},
  organization={Springer}
}

@article{ahmed2018survey,
  title={A survey on deep learning advances on different 3D data representations},
  author={Ahmed, Eman and Saint, Alexandre and Shabayek, Abd El Rahman and Cherenkova, Kseniya and Das, Rig and Gusev, Gleb and Aouada, Djamila and Ottersten, Bjorn},
  journal={arXiv preprint arXiv:1808.01462},
  year={2018}
}

@article{richman1975mechanical,
  title={Mechanical Model of Brain Convolutional Development: Pathologic and experimental data suggest a model based on differential growth within the cerebral cortex.},
  author={Richman, David P and Stewart, R Malcolm and Hutchinson, John and Caviness Jr, Verne S},
  journal={Science},
  volume={189},
  number={4196},
  pages={18--21},
  year={1975},
  publisher={American Association for the Advancement of Science}
}

@article{tallinen2016growth,
  title={On the growth and form of cortical convolutions},
  author={Tallinen, Tuomas and Chung, Jun Young and Rousseau, Fran{\c{c}}ois and Girard, Nadine and Lef{\`e}vre, Julien and Mahadevan, Lakshminarayanan},
  journal={Nature Physics},
  volume={12},
  number={6},
  pages={588--593},
  year={2016},
  publisher={Nature Publishing Group UK London}
}

@article{ronan2014differential,
  title={Differential tangential expansion as a mechanism for cortical gyrification},
  author={Ronan, Lisa and Voets, Natalie and Rua, Catarina and Alexander-Bloch, Aaron and Hough, Morgan and Mackay, Clare and Crow, Tim J and James, Anthony and Giedd, Jay N and Fletcher, Paul C},
  journal={Cerebral Cortex},
  volume={24},
  number={8},
  pages={2219--2228},
  year={2014},
  publisher={Oxford University Press}
}

@article{razavi2018genomic,
  title={The genomic landscape of endocrine-resistant advanced breast cancers},
  author={Razavi, Pedram and Chang, Matthew T and Xu, Guotai and Bandlamudi, Chaitanya and Ross, Dara S and Vasan, Neil and Cai, Yanyan and Bielski, Craig M and Donoghue, Mark TA and Jonsson, Philip and others},
  journal={Cancer cell},
  volume={34},
  number={3},
  pages={427--438},
  year={2018},
  publisher={Elsevier}
}

@article{bakhaty2017consistent,
  title={Consistent trilayer biomechanical modeling of aortic valve leaflet tissue},
  author={Bakhaty, Ahmed A and Govindjee, Sanjay and Mofrad, Mohammad RK},
  journal={Journal of Biomechanics},
  volume={61},
  pages={1--10},
  year={2017},
  publisher={Elsevier}
}

@article{madani2019bridging,
  title={Bridging finite element and machine learning modeling: stress prediction of arterial walls in atherosclerosis},
  author={Madani, Ali and Bakhaty, Ahmed and Kim, Jiwon and Mubarak, Yara and Mofrad, Mohammad RK},
  journal={Journal of biomechanical engineering},
  volume={141},
  number={8},
  pages={084502},
  year={2019},
  publisher={American Society of Mechanical Engineers}
}

@article{darayi2022computational,
  title={Computational models of cortical folding: a review of common approaches},
  author={Darayi, Mohsen and Hoffman, Mia E and Sayut, John and Wang, Shuolun and Demirci, Nagehan and Consolini, Jack and Holland, Maria A},
  journal={Journal of Biomechanics},
  volume={139},
  pages={110851},
  year={2022},
  publisher={Elsevier}
}

@article{barbeito2020modeling,
  title={Modeling the effect of brain growth on cranial bones using finite-element analysis and geometric morphometrics},
  author={Barbeito-Andr{\'e}s, Jimena and Bonfili, Noelia and Nogu{\'e}, Jordi Marc{\'e} and Bernal, Valeria and Gonzalez, Paula N},
  journal={Surgical and Radiologic Anatomy},
  volume={42},
  number={7},
  pages={741--748},
  year={2020},
  publisher={Springer}
}

@article{razavi2015role,
  title={Role of mechanical factors in cortical folding development},
  author={Razavi, Mir Jalil and Zhang, Tuo and Li, Xiao and Liu, Tianming and Wang, Xianqiao},
  journal={Physical Review E},
  volume={92},
  number={3},
  pages={032701},
  year={2015},
  publisher={APS}
}

@article{tallinen2015mechanics,
  title={Mechanics of invagination and folding: Hybridized instabilities when one soft tissue grows on another},
  author={Tallinen, Tuomas and Biggins, John S},
  journal={Physical Review E},
  volume={92},
  number={2},
  pages={022720},
  year={2015},
  publisher={APS}
}

@article{toro2005morphogenetic,
  title={A morphogenetic model for the development of cortical convolutions},
  author={Toro, Roberto and Burnod, Yves},
  journal={Cerebral cortex},
  volume={15},
  number={12},
  pages={1900--1913},
  year={2005},
  publisher={Oxford University Press}
}

@article{tallinen2014gyrification,
  title={Gyrification from constrained cortical expansion},
  author={Tallinen, Tuomas and Chung, Jun Young and Biggins, John S and Mahadevan, L},
  journal={Proceedings of the National Academy of Sciences},
  volume={111},
  number={35},
  pages={12667--12672},
  year={2014},
  publisher={National Acad Sciences}
}

@article{chavoshnejad2021role,
  title={Role of axonal fibers in the cortical folding patterns: A tale of variability and regularity},
  author={Chavoshnejad, Poorya and Li, Xiao and Zhang, Songyao and Dai, Weiying and Vasung, Lana and Liu, Tianming and Zhang, Tuo and Wang, Xianqiao and Razavi, Mir Jalil},
  journal={Brain Multiphysics},
  volume={2},
  pages={100029},
  year={2021},
  publisher={Elsevier}
}

@article{holland2015emerging,
  title={Emerging brain morphologies from axonal elongation},
  author={Holland, Maria A and Miller, Kyle E and Kuhl, Ellen},
  journal={Annals of biomedical engineering},
  volume={43},
  pages={1640--1653},
  year={2015},
  publisher={Springer}
}

@article{zhang2016mechanism,
  title={Mechanism of consistent gyrus formation: an experimental and computational study},
  author={Zhang, Tuo and Razavi, Mir Jalil and Li, Xiao and Chen, Hanbo and Liu, Tianming and Wang, Xianqiao},
  journal={Scientific reports},
  volume={6},
  number={1},
  pages={37272},
  year={2016},
  publisher={Nature Publishing Group UK London}
}

@article{razavi2017radial,
  title={Radial structure scaffolds convolution patterns of developing cerebral cortex},
  author={Razavi, Mir Jalil and Zhang, Tuo and Chen, Hanbo and Li, Yujie and Platt, Simon and Zhao, Yu and Guo, Lei and Hu, Xiaoping and Wang, Xianqiao and Liu, Tianming},
  journal={Frontiers in computational neuroscience},
  volume={11},
  pages={76},
  year={2017},
  publisher={Frontiers Media SA}
}

@article{thompson1996surface,
  title={A surface-based technique for warping three-dimensional images of the brain},
  author={Thompson, Paul and Toga, Arthur W},
  journal={IEEE transactions on medical imaging},
  volume={15},
  number={4},
  pages={402--417},
  year={1996},
  publisher={IEEE}
}

@article{van1998functional,
  title={Functional and structural mapping of human cerebral cortex: solutions are in the surfaces},
  author={Van Essen, David C and Drury, Heather A and Joshi, Sarang and Miller, Michael I},
  journal={Proceedings of the National Academy of Sciences},
  volume={95},
  number={3},
  pages={788--795},
  year={1998},
  publisher={National Acad Sciences}
}

@article{woods1998automated,
  title={Automated image registration: II. Intersubject validation of linear and nonlinear models},
  author={Woods, Roger P and Grafton, Scott T and Watson, John DG and Sicotte, Nancy L and Mazziotta, John C},
  journal={Journal of computer assisted tomography},
  volume={22},
  number={1},
  pages={153--165},
  year={1998},
  publisher={LWW}
}

@article{fischl1999cortical,
  title={Cortical surface-based analysis: II: inflation, flattening, and a surface-based coordinate system},
  author={Fischl, Bruce and Sereno, Martin I and Dale, Anders M},
  journal={Neuroimage},
  volume={9},
  number={2},
  pages={195--207},
  year={1999},
  publisher={Elsevier}
}

@article{fischl2002whole,
  title={Whole brain segmentation: automated labeling of neuroanatomical structures in the human brain},
  author={Fischl, Bruce and Salat, David H and Busa, Evelina and Albert, Marilyn and Dieterich, Megan and Haselgrove, Christian and Van Der Kouwe, Andre and Killiany, Ron and Kennedy, David and Klaveness, Shuna and others},
  journal={Neuron},
  volume={33},
  number={3},
  pages={341--355},
  year={2002},
  publisher={Elsevier}
}

@article{mangin2004coordinate,
  title={Coordinate-based versus structural approaches to brain image analysis},
  author={Mangin, J-F and Rivi{\`e}re, Denis and Coulon, Olivier and Poupon, Cyril and Cachia, Arnaud and Cointepas, Yann and Poline, J-B and Le Bihan, Denis and R{\'e}gis, Jean and Papadopoulos-Orfanos, Dimitri},
  journal={Artificial intelligence in Medicine},
  volume={30},
  number={2},
  pages={177--197},
  year={2004},
  publisher={Elsevier}
}

@inproceedings{li2008novel,
  title={A novel method for cortical sulcal fundi extraction},
  author={Li, Gang and Liu, Tianming and Nie, Jingxin and Guo, Lei and Wong, Stephen TC},
  booktitle={Medical Image Computing and Computer-Assisted Intervention--MICCAI 2008: 11th International Conference, New York, NY, USA, September 6-10, 2008, Proceedings, Part I 11},
  pages={270--278},
  year={2008},
  organization={Springer}
}

@inproceedings{yang2008diffusion,
  title={Diffusion tensor image registration using tensor geometry and orientation features},
  author={Yang, Jinzhong and Shen, Dinggang and Davatzikos, Christos and Verma, Ragini},
  booktitle={International Conference on Medical Image Computing and Computer-Assisted Intervention},
  pages={905--913},
  year={2008},
  organization={Springer}
}

@inproceedings{awate2009multivariate,
  title={Multivariate high-dimensional cortical folding analysis, combining complexity and shape, in neonates with congenital heart disease},
  author={Awate, Suyash P and Yushkevich, Paul and Song, Zhuang and Licht, Daniel and Gee, James C},
  booktitle={Information Processing in Medical Imaging: 21st International Conference, IPMI 2009, Williamsburg, VA, USA, July 5-10, 2009. Proceedings 21},
  pages={552--563},
  year={2009},
  organization={Springer}
}

@inproceedings{boucher2009oriented,
  title={Oriented morphometry of folds on surfaces},
  author={Boucher, Maxime and Evans, Alan and Siddiqi, Kaleem},
  booktitle={Information Processing in Medical Imaging: 21st International Conference, IPMI 2009, Williamsburg, VA, USA, July 5-10, 2009. Proceedings 21},
  pages={614--625},
  year={2009},
  organization={Springer}
}

@article{li2010gyral,
  title={Gyral folding pattern analysis via surface profiling},
  author={Li, Kaiming and Guo, Lei and Li, Gang and Nie, Jingxin and Faraco, Carlos and Cui, Guangbin and Zhao, Qun and Miller, L Stephen and Liu, Tianming},
  journal={NeuroImage},
  volume={52},
  number={4},
  pages={1202--1214},
  year={2010},
  publisher={Elsevier}
}

@article{fernandez2016cerebral,
  title={Cerebral cortex expansion and folding: what have we learned?},
  author={Fern{\'a}ndez, Virginia and Llinares-Benadero, Cristina and Borrell, V{\'\i}ctor},
  journal={The EMBO journal},
  volume={35},
  number={10},
  pages={1021--1044},
  year={2016}
}

@inproceedings{wang2019early,
  title={On early brain folding patterns using biomechanical growth modeling},
  author={Wang, Xiaoyu and Bohi, Amine and Al Harrach, Mariam and Dinomais, Mickael and Lef{\`e}vre, Julien and Rousseau, Fran{\c{c}}ois},
  booktitle={2019 41st Annual International Conference of the IEEE Engineering in Medicine and Biology Society (EMBC)},
  pages={146--149},
  year={2019},
  organization={IEEE}
}

@article{da2021role,
  title={The role of thickness inhomogeneities in hierarchical cortical folding},
  author={da Costa Campos, Lucas and Hornung, Raphael and Gompper, Gerhard and Elgeti, Jens and Caspers, Svenja},
  journal={NeuroImage},
  volume={231},
  pages={117779},
  year={2021},
  publisher={Elsevier}
}

@article{garcia2021model,
  title={A model of tension-induced fiber growth predicts white matter organization during brain folding},
  author={Garcia, Kara E and Wang, Xiaojie and Kroenke, Christopher D},
  journal={Nature communications},
  volume={12},
  number={1},
  pages={6681},
  year={2021},
  publisher={Nature Publishing Group UK London}
}

@article{zarzor2021two,
  title={A two-field computational model couples cellular brain development with cortical folding},
  author={Zarzor, MS and Kaessmair, S and Steinmann, P and Bl{\"u}mcke, I and Budday, S},
  journal={Brain Multiphysics},
  volume={2},
  pages={100025},
  year={2021},
  publisher={Elsevier}
}

@misc{chang2015shapenetinformationrich3dmodel,
      title={ShapeNet: An Information-Rich 3D Model Repository}, 
      author={Angel X. Chang and Thomas Funkhouser and Leonidas Guibas and Pat Hanrahan and Qixing Huang and Zimo Li and Silvio Savarese and Manolis Savva and Shuran Song and Hao Su and Jianxiong Xiao and Li Yi and Fisher Yu},
      year={2015},
      eprint={1512.03012},
      archivePrefix={arXiv},
      primaryClass={cs.GR},
      url={https://arxiv.org/abs/1512.03012}, 
}

@inproceedings{pan2021variational,
  title={Variational relational point completion network},
  author={Pan, Liang and Chen, Xinyi and Cai, Zhongang and Zhang, Junzhe and Zhao, Haiyu and Yi, Shuai and Liu, Ziwei},
  booktitle={Proceedings of the IEEE/CVF conference on computer vision and pattern recognition},
  pages={8524--8533},
  year={2021}
}

@misc{wu20153dshapenetsdeeprepresentation,
      title={3D ShapeNets: A Deep Representation for Volumetric Shapes}, 
      author={Zhirong Wu and Shuran Song and Aditya Khosla and Fisher Yu and Linguang Zhang and Xiaoou Tang and Jianxiong Xiao},
      year={2015},
      eprint={1406.5670},
      archivePrefix={arXiv},
      primaryClass={cs.CV},
      url={https://arxiv.org/abs/1406.5670}, 
}

@misc{xu2018spidercnndeeplearningpoint,
      title={SpiderCNN: Deep Learning on Point Sets with Parameterized Convolutional Filters}, 
      author={Yifan Xu and Tianqi Fan and Mingye Xu and Long Zeng and Yu Qiao},
      year={2018},
      eprint={1803.11527},
      archivePrefix={arXiv},
      primaryClass={cs.CV},
      url={https://arxiv.org/abs/1803.11527}, 
}

@inproceedings{bourbia2022blind,
  title={Blind Projection-based 3D Point Cloud Quality Assessment Method using a Convolutional Neural Network.},
  author={Bourbia, Salima and Karine, Ayoub and Chetouani, Aladine and El Hassouni, Mohammed},
  booktitle={VISIGRAPP (4: VISAPP)},
  pages={518--525},
  year={2022}
}

@inproceedings{fan2017point,
  title={A point set generation network for 3d object reconstruction from a single image},
  author={Fan, Haoqiang and Su, Hao and Guibas, Leonidas J},
  booktitle={Proceedings of the IEEE conference on computer vision and pattern recognition},
  pages={605--613},
  year={2017}
}

@article{lu20223dctn,
  title={3DCTN: 3D convolution-transformer network for point cloud classification},
  author={Lu, Dening and Xie, Qian and Gao, Kyle and Xu, Linlin and Li, Jonathan},
  journal={IEEE Transactions on Intelligent Transportation Systems},
  volume={23},
  number={12},
  pages={24854--24865},
  year={2022},
  publisher={IEEE}
}

@article{vaswani2017attention,
  title={Attention is all you need},
  author={Vaswani, A},
  journal={Advances in Neural Information Processing Systems},
  year={2017}
}

@inproceedings{yang2019modeling,
  title={Modeling point clouds with self-attention and gumbel subset sampling},
  author={Yang, Jiancheng and Zhang, Qiang and Ni, Bingbing and Li, Linguo and Liu, Jinxian and Zhou, Mengdie and Tian, Qi},
  booktitle={Proceedings of the IEEE/CVF conference on computer vision and pattern recognition},
  pages={3323--3332},
  year={2019}
}

@article{ioffe2015batch,
  title={Batch normalization: Accelerating deep network training by reducing internal covariate shift},
  author={Ioffe, Sergey},
  journal={arXiv preprint arXiv:1502.03167},
  year={2015}
}

@article{zhu2017face,
  title={Face alignment in full pose range: A 3d total solution},
  author={Zhu, Xiangyu and Liu, Xiaoming and Lei, Zhen and Li, Stan Z},
  journal={IEEE transactions on pattern analysis and machine intelligence},
  volume={41},
  number={1},
  pages={78--92},
  year={2017},
  publisher={IEEE}
}

@inproceedings{he2016deep,
  title={Deep residual learning for image recognition},
  author={He, Kaiming and Zhang, Xiangyu and Ren, Shaoqing and Sun, Jian},
  booktitle={Proceedings of the IEEE conference on computer vision and pattern recognition},
  pages={770--778},
  year={2016}
}

@article{zhou2020batch,
  title={Batch group normalization},
  author={Zhou, Xiao-Yun and Sun, Jiacheng and Ye, Nanyang and Lan, Xu and Luo, Qijun and Lai, Bo-Lin and Esperanca, Pedro and Yang, Guang-Zhong and Li, Zhenguo},
  journal={arXiv preprint arXiv:2012.02782},
  year={2020}
}

@misc{simonyan2015deepconvolutionalnetworkslargescale,
      title={Very Deep Convolutional Networks for Large-Scale Image Recognition}, 
      author={Karen Simonyan and Andrew Zisserman},
      year={2015},
      eprint={1409.1556},
      archivePrefix={arXiv},
      primaryClass={cs.CV},
      url={https://arxiv.org/abs/1409.1556}, 
}

@INPROCEEDINGS{8578839,
  author={Deng, Jiankang and Cheng, Shiyang and Xue, Niannan and Zhou, Yuxiang and Zafeiriou, Stefanos},
  booktitle={2018 IEEE/CVF Conference on Computer Vision and Pattern Recognition}, 
  title={UV-GAN: Adversarial Facial UV Map Completion for Pose-Invariant Face Recognition}, 
  year={2018},
  volume={},
  number={},
  pages={7093-7102},
  doi={10.1109/CVPR.2018.00741}}

@article{Na_2020,
   title={Facial UV map completion for pose-invariant face recognition: a novel adversarial approach based on coupled attention residual UNets},
   volume={10},
   ISSN={2192-1962},
   url={http://dx.doi.org/10.1186/s13673-020-00250-w},
   DOI={10.1186/s13673-020-00250-w},
   number={1},
   journal={Human-centric Computing and Information Sciences},
   publisher={Springer Science and Business Media LLC},
   author={Na, In Seop and Tran, Chung and Nguyen, Dung and Dinh, Sang},
   year={2020},
   month=nov }

@article{article,
author = {Heckbert, Paul},
year = {1986},
month = {12},
pages = {56-67},
title = {Survey Of Texture Mapping},
volume = {6},
journal = {Computer Graphics and Applications, IEEE},
doi = {10.1109/MCG.1986.276672}
}

@article{RUDIN1992259,
title = {Nonlinear total variation based noise removal algorithms},
journal = {Physica D: Nonlinear Phenomena},
volume = {60},
number = {1},
pages = {259-268},
year = {1992},
issn = {0167-2789},
doi = {https://doi.org/10.1016/0167-2789(92)90242-F},
url = {https://www.sciencedirect.com/science/article/pii/016727899290242F},
author = {Leonid I. Rudin and Stanley Osher and Emad Fatemi}
}

@misc{johnson2016perceptuallossesrealtimestyle,
      title={Perceptual Losses for Real-Time Style Transfer and Super-Resolution}, 
      author={Justin Johnson and Alexandre Alahi and Li Fei-Fei},
      year={2016},
      eprint={1603.08155},
      archivePrefix={arXiv},
      primaryClass={cs.CV},
      url={https://arxiv.org/abs/1603.08155}, 
}

@misc{gatys2015neuralalgorithmartisticstyle,
      title={A Neural Algorithm of Artistic Style}, 
      author={Leon A. Gatys and Alexander S. Ecker and Matthias Bethge},
      year={2015},
      eprint={1508.06576},
      archivePrefix={arXiv},
      primaryClass={cs.CV},
      url={https://arxiv.org/abs/1508.06576}, 
}

@article{ge2018denser,
  title={Denser growing fiber connections induce 3-hinge gyral folding},
  author={Ge, Fangfei and Li, Xiao and Razavi, Mir Jalil and Chen, Hanbo and Zhang, Tuo and Zhang, Shu and Guo, Lei and Hu, Xiaoping and Wang, Xianqiao and Liu, Tianming},
  journal={Cerebral Cortex},
  volume={28},
  number={3},
  pages={1064--1075},
  year={2018},
  publisher={Oxford University Press}
}

@article{razavi2021mechanism,
  title={Mechanism exploration of 3-hinge gyral formation and pattern recognition},
  author={Razavi, Mir Jalil and Liu, Tianming and Wang, Xianqiao},
  journal={Cerebral Cortex Communications},
  volume={2},
  number={3},
  pages={tgab044},
  year={2021},
  publisher={Oxford University Press}
}

@article{cao2012buckling,
  title={Buckling and post-buckling of a stiff film resting on an elastic graded substrate},
  author={Cao, Yan-Ping and Jia, Fei and Zhao, Yan and Feng, Xi-Qiao and Yu, Shou-Wen},
  journal={International Journal of Solids and Structures},
  volume={49},
  number={13},
  pages={1656--1664},
  year={2012},
  publisher={Elsevier}
}

@article{solhtalab2025stress,
  title={Stress landscape of folding brain serves as a map for axonal pathfinding},
  author={Solhtalab, Akbar and Foroughi, Ali H and Pierotich, Lana and Razavi, Mir Jalil},
  journal={Nature Communications},
  volume={16},
  number={1},
  pages={1187},
  year={2025},
  publisher={Nature Publishing Group UK London}
}

@article{solhtalab2025mechanics,
  title={Mechanics of the Spatiotemporal Evolution of Sulcal Pits in the Folding Brain},
  author={Solhtalab, Akbar and Guo, Yanchen and Gholipour, Ali and Dai, Weiying and Razavi, Mir Jalil},
  journal={Human Brain Mapping},
  volume={46},
  number={13},
  pages={e70332},
  year={2025},
  publisher={Wiley Online Library}
}

@article{budday2014role,
  title={The role of mechanics during brain development},
  author={Budday, Silvia and Steinmann, Paul and Kuhl, Ellen},
  journal={Journal of the Mechanics and Physics of Solids},
  volume={72},
  pages={75--92},
  year={2014},
  publisher={Elsevier}
}

@article{chavoshnejad2024theoretical,
  title={A theoretical framework for predicting the heterogeneous stiffness map of brain white matter tissue},
  author={Chavoshnejad, Poorya and Li, Guangfa and Solhtalab, Akbar and Liu, Dehao and Razavi, Mir Jalil},
  journal={Physical Biology},
  volume={21},
  number={6},
  pages={066004},
  year={2024},
  publisher={IOP Publishing}
}



\appendix

\end{document}